\pdfoutput=1

\documentclass[11pt]{article}

\usepackage[final]{acl}

\usepackage{times}
\usepackage{latexsym} 

\usepackage[T1]{fontenc}

\usepackage[utf8]{inputenc}

\usepackage{microtype}

\usepackage{inconsolata}

\usepackage{graphicx}

\usepackage{tabularx}
\usepackage{xcolor}
\usepackage{hyperref}
\usepackage{xurl}

\usepackage{paralist}
\usepackage{tcolorbox}
\tcbuselibrary{breakable}

\usepackage{booktabs}
\usepackage{multirow}
\usepackage{subfigure}
\usepackage{wrapfig}
\usepackage{caption}
\usepackage{amsmath,amssymb}
\usepackage{colortbl}
\usepackage{arydshln}

\usepackage{fvextra} 
\DefineVerbatimEnvironment{VerbatimWrap}{Verbatim}{
breaklines=true, 
breakindent=0pt, 
breaksymbol={},
fontsize=\small,
}

\definecolor{darkblue}{rgb}{0, 0, 0.5}
\definecolor{chocolate}{HTML}{D2691E}
\definecolor{maroon}{HTML}{A00000}
\definecolor{indigo}{HTML}{4B0082}

\usepackage{cleveref}
\crefformat{section}{\S#2#1#3} 
\crefformat{subsection}{\S#2#1#3}
\crefformat{subsubsection}{\S#2#1#3}

\title{Open-Domain Safety Policy Construction}

{\centering
    \author{Di Wu\textsuperscript{\textnormal{1}}, Siyue Liu\textsuperscript{\textnormal{1}}, Zixiang Ji\textsuperscript{\textnormal{1}}, \textbf{Ya-Liang Chang}\textsuperscript{\textnormal{2}}, \textbf{Zhe-Yu Liu}\textsuperscript{\textnormal{2}},\\\textbf{Andrew Pleffer}\textsuperscript{\textnormal{2}}, \textbf{Kai-Wei Chang\textsuperscript{\textnormal{1}}}
        \\
    \textsuperscript{\textnormal{1}}University of California, Los Angeles\ \ \ \   \textsuperscript{\textnormal{2}}Taboola \\
    \texttt{\{diwu, kwchang\}@cs.ucla.edu} \\
}
}

\begin{document}
\maketitle

\begin{abstract}
Moderation layers are increasingly a core component of many products built on user- or model-generated content. However, drafting and maintaining domain-specific safety policies remains costly. We present \textbf{Deep Policy Research} (DPR), a minimal agentic system that drafts a full content moderation policy based on only human-written seed domain information. DPR uses a single web search tool and lightweight scaffolding to iteratively propose search queries, distill diverse web sources into policy rules, and organize rules into an indexed document. We evaluate DPR on (1) the OpenAI undesired content benchmark across five domains with two compact reader LLMs and (2) an in-house multimodal advertisement moderation benchmark. DPR consistently outperforms definition-only and in-context learning baselines, and in our end-to-end setting it is competitive with expert-written policy sections in several domains. Moreover, under the same seed specification and evaluation protocol, DPR outperforms a general-purpose deep research system, suggesting that a task-specific, structured research loop can be more effective than generic web research for policy drafting. We release our experiment code at  \url{https://github.com/xiaowu0162/deep-policy-research}.
\end{abstract}
\section{Introduction}

Content moderation modules are core layers in modern products for managing unsafe or low-quality inputs. These systems are guided by domain-specific policies that define allowed and disallowed content and support consistent enforcement across labeling, training, and deployment \citep{DBLP:conf/aaai/MarkovZANLAJW23, vidgen2020directions, yin2021towards, Zeng2020BadNews}. However, drafting and maintaining high-quality policies remains costly. It requires domain expertise, repeated iteration, and frequent updates as products evolve and new edge cases appear. While studies have explored automatic pipelines to improve the effectiveness or reduce the cost of applying the policies \citet{DBLP:conf/aaai/MarkovZANLAJW23, DBLP:journals/corr/abs-2412-16339}, a fully human-written policy is still a prerequisite. In this paper, we challenge this assumption by asking: 

\begin{center}
    \textit{Can we leverage LLMs to assist in drafting the policies themselves?}
\end{center}

\begin{figure}[t]
    \centering
    \includegraphics[scale=0.2]{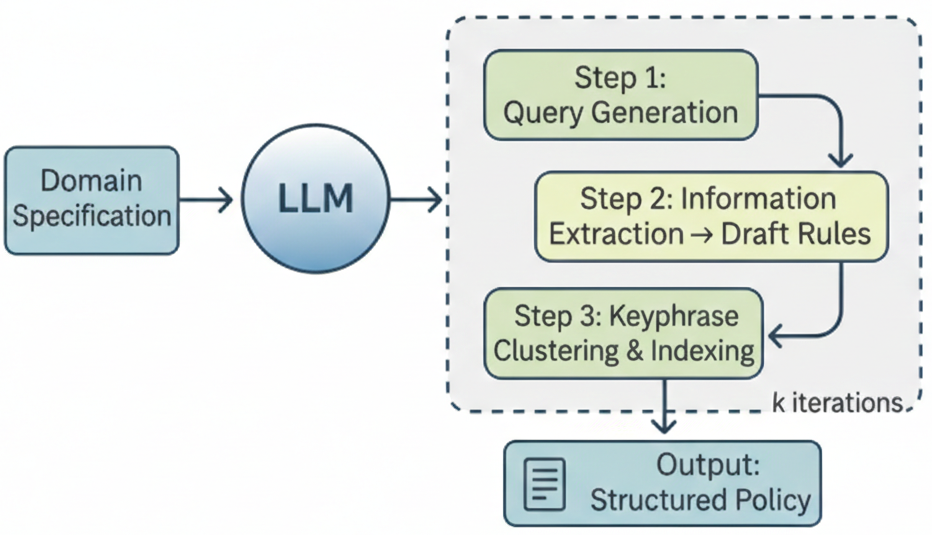}
    \caption{An illustration of Deep Policy Research. Based on a domain specification, an LLM iteratively interacts with a search engine, extracts policy rules, and indexes the rules through keyphrase-based clustering.}
    \label{fig:main-figure}
    \vspace{-3mm}
\end{figure}

To begin with, we frame the task \textbf{open-domain policy construction}. The input is a concise domain specification and access to a search engine. The output is a structured policy document. Success is measured by downstream utility, such as the accuracy of a fixed moderation model when the policy is provided in-context.

We then propose \textbf{Deep Policy Research} (DPR). DPR is a minimal agent that uses only web search as an external tool and a lightweight scaffolding scheme for rule writing. Starting from a one-sentence domain definition, DPR iteratively identifies missing coverage, issues targeted queries, distills retrieved sources into rule predicates, and consolidates them into an indexed policy document (\Cref{fig:main-figure}). The indexing stage organizes rules into coherent sections, improving readability and helping reader models consume long policies.

We evaluate DPR in two content moderation settings. On the OpenAI undesired content benchmark \citep{DBLP:conf/aaai/MarkovZANLAJW23} across five domains, DPR improves moderation F$_1$ over both definition-only prompting and few-shot in-context examples for two compact reader LLMs. Averaged across domains, DPR increases F$_1$ from 0.752 to 0.792 on Llama 3.1 8B and from 0.810 to 0.831 on Qwen2.5 7B, with the largest gains on more subjective categories such as Violence, Harassment, and Self-Harm (\Cref{section-acc-results-openai}). Under the same seed specification and evaluation protocol, DPR also outperforms a general-purpose deep research system, improving average F$_1$ from 0.776 to 0.792 on Llama 3.1 8B and from 0.800 to 0.831 on Qwen2.5 7B (\Cref{section-acc-results-openai}). We further evaluate DPR on an in-house multimodal advertisement moderation benchmark. Replacing an expert-written domain section with a DPR-generated section recovers much of the human policy benefit in several domains and substantially improves over removing the section or using only the one-sentence specification (\Cref{section-acc-results-taboola}).

In summary, we introduce open-domain policy construction and evaluate it by downstream utility. We propose Deep Policy Research, a minimal agent that uses web search and lightweight scaffolding to synthesize and index domain-specific policies. Across text-only and multimodal moderation settings, DPR-generated policies improve downstream moderation and are competitive with expert-written policy sections in several domains, while outperforming a general-purpose deep research baseline under the same protocol. DPR also provides a reproducible environment for future work on policy drafting agents, including reader-model-specific policy presentation and generating illustrative examples alongside rule predicates to clarify decision boundaries.

\section{Related Work}

\paragraph{Policy Use in LLM Systems} Recent alignment strategies explicitly incorporate human-written safety policies into the training or reasoning process of large language models. Deliberative Alignment fine-tunes LLMs to reason based on an entire written policy before responding, yielding safer outputs that strictly adhere to guidelines (e.g. reduced jailbreaks and fewer unjustified refusals) \cite{DBLP:journals/corr/abs-2412-16339}. Anthropic’s Constitutional AI similarly forgoes direct human feedback in favor of a fixed set of normative principles that the model internalizes as a “constitution,” using them to self-criticize and refine its answers \cite{bai2022constitutional}. Other works integrate policies as part of the reward or decision mechanism: for instance, OpenAI’s GPT-4 alignment process included a rule-based reward model that penalized policy violations during RLHF \cite{mu2024rule}, and recent safety reasoning frameworks train models to follow chain-of-thought traces grounded in explicit policy rules \cite{mou2025saro}.

\paragraph{Policy Writing} The direction of automatically creating or refining safety policies has been less explored. OpenAI has demonstrated that GPT-4 can assist policy designers by identifying ambiguities and edge cases in draft guidelines: the model labels content according to a given policy, explains any discrepancies with human judgments, and suggests clarifications, thereby accelerating the policy refinement loop \cite{openai2023moderation}. We build upon this intuition and take a step forward, automatically generating policy end-to-end from human-curated domain specifications.

\section{Approach}
\label{section-approach}

\subsection{Problem Formulation}
\label{section-problem-formulation}

We study \textbf{open-domain policy construction}. The input is a domain specification $s$ that describes scope and intent for a single moderation domain, and a search engine $\mathcal{G}$. The output is a policy document $P$ consisting of a set of textual rules organized into sections. $P$ can be free-formed or following a hierarchical structure. We evaluate $P$ by \textbf{downstream utility} in a fixed content moderation setup, where a reader LLM receives $P$ in-context and performs safe vs. unsafe binary classifications. A policy construction system succeeds if its output policy improves downstream moderation performance under the same reader model and evaluation protocol.

\subsection{Deep Policy Research}
\label{section-dpr-system-intro}

We present \textbf{Deep Policy Research} (DPR), a minimal research agent that constructs a policy by iteratively searching the web and distilling sources into structured rules. DPR uses an LLM $\mathcal{M}$ as the research model and web search $\mathcal{G}$ as the only external tool. DPR runs for $k$ iterations and maintains two artifacts at iteration $i$: a policy draft $P_i$ and an index $I_i$ that organizes rules into sections. DPR initializes $P_0 \equiv s$ and $I_0 \equiv s$.

Concretely, at each iteration $i \in \{1,\dots,k\}$, DPR performs three steps.

\paragraph{Step 1: Query generation.}
DPR first analyzes the current policy organization $I_{i-1}$ and proposes a set of research queries $Q_i$ to expand coverage or refine ambiguous parts of the policy. Queries are written to target definitional boundaries, common edge cases, high-risk subtypes, and enforcement cues. For each query $q \in Q_i$, DPR retrieves the top $m$ search results using $\mathcal{G}$ and collects page titles, snippets, and URLs as evidence for rule extraction.

\paragraph{Step 2: Rule extraction and consolidation.}
Given the retrieved evidence, DPR prompts $\mathcal{M}$ to extract candidate rules in a consistent schema. Each rule is written as a short predicate-style statement with a clear decision boundary and optional qualifiers. Concretely, we ask $\mathcal{M}$ to produce rules that specify a condition and a moderation decision, and to include brief scope qualifiers when needed.
DPR then runs a self-critique pass to improve precision and reduce noise. In this pass, $\mathcal{M}$ removes irrelevant or overly generic rules, merges redundant rules that express the same decision boundary, and resolves conflicts by preferring rules that are supported by multiple sources or by higher-quality sources. The output is a consolidated rule set $R_i$.

\paragraph{Step 3: Indexing.}
DPR merges the new rules into the policy draft, $P_i \leftarrow P_{i-1} \cup R_i$. It then organizes the full rule set into sections to form an indexed policy document $I_i$. We use keyphrase-based clustering to build this index. DPR asks $\mathcal{M}$ to extract keyphrases for each rule, clusters keyphrases into $n$ groups using k-means, and asks $\mathcal{M}$ to name each cluster and write a short section summary that captures the shared theme. Finally, DPR merges clusters with overlapping semantics to produce a compact, readable index. The resulting $I_i$ is used both as a human-readable policy document and as a coverage signal for the next query generation step.

After $k$ iterations, DPR outputs the final indexed policy $P \equiv I_k$. \Cref{fig:main-figure} illustrates the overall loop. Our goal is not to engineer complex agent architectures or elaborate scaffolding strategies. Instead, we aim to isolate a minimal, reproducible research loop that uses a single external tool and lightweight human scaffolding, and to show that this simple design can already produce useful, structured policies with measurable downstream utility. For reproduction, we list the domain specifications in \Cref{section-dpr-detail-domain-specifications} and the prompts in \Cref{section-dpr-detail-prompts}.

\begin{table*}[h]
\centering
\renewcommand{\arraystretch}{1.3} 
\resizebox{0.7\linewidth}{!}{
\begin{tabular}{l|ccccc|c}
\hline 
\textbf{} & \textbf{Sexual} & \textbf{Hate} & \textbf{Violence} & \textbf{Harassment} & \textbf{Self-Harm} & \textbf{Average} \\
\hline 

\rowcolor{gray!20} \multicolumn{7}{c}{\textbf{\texttt{Llama 3.1 8B Instruct}}} \\
\hline 
Seed Information & 0.916 & 0.757 & 0.658 & 0.640 & 0.788 & 0.752 \\
In-Context Learning & 0.923 & 0.728 & 0.582 & 0.610 & 0.691 & 0.707 \\
OAI DR & 0.828 & 0.800 & 0.738 & 0.683 & 0.829 & 0.776 \\
DPR & 0.910 & 0.813 & 0.717 & 0.683 & 0.835 & \textbf{0.792} \\
\hline 

\rowcolor{gray!20} \multicolumn{7}{c}{\textbf{\texttt{Qwen2.5 7B Instruct}}} \\
\hline 
Seed Information & 0.939 & 0.817 & 0.782 & 0.670 & 0.842 & 0.810 \\
In-Context Learning & 0.927 & 0.803 & 0.645 & 0.551 & 0.779 & 0.741 \\
OAI DR & 0.919 & 0.765 & 0.763 & 0.781 & 0.773 & 0.800 \\
DPR & 0.949 & 0.811 & 0.784 & 0.752 & 0.860 & \textbf{0.831} \\
\hline 
\end{tabular}
}
\caption{Evaluation results on OpenAI Content Moderation. DPR improves content moderation F1, outperforming both only using seed information or performing in-context learning with human-written examples.}
\label{tab:safety-performance-openai}
\end{table*}

\section{Results: OpenAI Content Moderation}
\label{section-results-openai}

\subsection{Experimental Setup}
\label{section-eval-details-openai}

We evaluate DPR on the OpenAI undesired content benchmark introduced in \citet{DBLP:conf/aaai/MarkovZANLAJW23}. Following prior work, we consider five major moderation domains and report binary classification F$_1$ computed per domain and averaged across domains. In all settings, we keep the downstream reader LLM fixed and vary only the policy provided in-context, so differences reflect the utility of the constructed policy rather than changes in the classifier or model weights. Full implementation details are provided in \Cref{section-further-impl-details-openai}.

\paragraph{Baselines} We compare DPR with three baselines:
\begin{compactitem}
    \item \textbf{Seed Information}: judging with only the seed information $s_i$ for each domain. 
    \item \textbf{In-Context Learning}: we randomly sample three unsafe examples and three safe examples as the in-context demonstrations. 
    \item \textbf{OAI DR}: we manually run the OpenAI Deep Research Agent through the WebUI. The seed information is provided and we use GPT-5.1 as the research model. 
\end{compactitem}

\subsection{Content Moderation Accuracy}
\label{section-acc-results-openai}

\Cref{tab:safety-performance-openai} reports results across five domains and two reader LLMs. DPR consistently improves over both Seed Information and In-Context Learning for every domain and reader model. Averaged across domains, DPR increases F$_1$ from 0.752 to 0.792 on Llama 3.1 8B and from 0.810 to 0.831 on Qwen2.5 7B. Gains are most pronounced in more subjective categories such as Violence, Harassment, and Self-Harm, where definition-only prompts leave substantial ambiguity and few-shot demonstrations are brittle. Notably, DPR does not introduce regressions on well-specified categories. For Sexual, differences are within 0.01 F$_1$ for both readers, indicating that web-grounded refinement can add coverage without injecting noise.

Under the same seed specification and evaluation protocol, DPR also outperforms the general-purpose deep research baseline. On Llama 3.1 8B, the average F$_1$ improves from 0.776 with OAI DR to 0.792 with DPR. On Qwen2.5 7B, the average improves from 0.800 to 0.831. This suggests that task-specific structure matters for policy drafting: a simple loop that enforces rule extraction, consolidation, and indexed organization can yield more usable policies than generic web research when the end goal is downstream moderation.

Overall, these results support two conclusions. First, open-domain policy construction can translate into measurable moderation gains even when the downstream model is held constant. Second, a minimal agent with a single external tool can be effective when paired with lightweight human scaffolding that constrains outputs into actionable rule predicates and an indexed document. We provide additional analyses on the research model and indexing design in \Cref{section-analyses-openai}. Further qualitative examples are provided in \Cref{section-qual-study}.

\section{Results: In-house Multimodal Advertisement Moderation}
\label{section-results-taboola}

\subsection{Experimental Setup}
\label{section-eval-details-taboola}

We further evaluate DPR in a real-world multimodal moderation setting using an in-house advertisement benchmark. Each example consists of ad text and a thumbnail image, and the task is to predict whether the creative complies with an in-house safety policy. We report binary classification F$_1$ and evaluate with a fixed vision-language reader model that receives the policy in-context. We consider two inference settings, single-sample decoding (S.S.) and majority voting (M.V.) over multiple samples. Full dataset and evaluation details are provided in \Cref{section-further-impl-details-taboola}.

This benchmark includes a comprehensive human-written policy document, where each section corresponds to a policy domain. Our evaluation therefore focuses on \textbf{substituting} a single domain section while keeping the remainder of the policy fixed. We compare five configurations:
\begin{compactenum}
    \item \textbf{No Policy} removes the target domain section entirely while keeping the other sections. 
    \item \textbf{Human Policy} uses the full original expert-written policy document. 
    \item \textbf{Seed Information} replaces the section with the one-sentence domain specification. 
    \item \textbf{DPR} replaces the section with a DPR-generated policy built from the same one-sentence specification and web search. 
    \item \textbf{DPR + Summary} further compresses the DPR index into a shorter rule set to reduce prompt length.

\end{compactenum}

\subsection{Content Moderation Accuracy}
\label{section-acc-results-taboola}

\Cref{tab:safety-performance-taboola} reports results for four domains under both inference settings. Removing the domain section or replacing it with only the one-sentence specification degrades performance in three of four domains, showing that the domain-specific section carries substantial utility beyond the rest of the policy. In contrast, substituting the section with DPR recovers much of this benefit. Under single-sample inference, DPR improves the average F$_1$ from 0.68 with No Policy and 0.69 with Seed Information to 0.75. Gains are largest for visually nuanced categories such as Exploitative and Offensive, where the web-sourced rules often capture concrete cues and common edge cases that are missing from a short specification. 

\begin{table*}[t!]
\centering
\renewcommand{\arraystretch}{1.3} 
\resizebox{0.7\linewidth}{!}{
\begin{tabular}{l|cc|cc|cc|cc}
\hline 
 & 
\multicolumn{2}{c|}{\textbf{Misrepresentative}} & 
\multicolumn{2}{c|}{\textbf{Finance Claims}} & 
\multicolumn{2}{c|}{\textbf{Exploitative}} & 
\multicolumn{2}{c}{\textbf{Offensive}} \\
& \textbf{S.S.} & \textbf{M.V.} 
& \textbf{S.S.} & \textbf{M.V.} 
& \textbf{S.S.} & \textbf{M.V.} 
& \textbf{S.S.} & \textbf{M.V.} \\
\hline 
No Policy & 0.714 & 0.727 & 0.500 & 0.509 & 0.793 & 0.782 & 0.714 & 0.786 \\
Seed Information & 0.701 & 0.701 & 0.553 & 0.544 & 0.839 & \textbf{0.885} & 0.679 & 0.786 \\
Human Policy & \textbf{0.740} & \textbf{0.779} & \textbf{0.833} & \textbf{0.877} & \textbf{0.920} & 0.908 & \textbf{0.893} & \textbf{0.964} \\
DPR & 0.727 & 0.740 & 0.597 & 0.597 & 0.908 & \textbf{0.920} & 0.786 & 0.821 \\
DPR + Summary & 0.701 & \textbf{0.779} & 0.588 & 0.614 & 0.874 & 0.908 & 0.821 & 0.821 \\
\hline 
\end{tabular}
}
\caption{Evaluation results on in-house multimodal moderation data. DPR improves content moderation F1, outperforming both no policy and only using seed information. DPR also performs on par with human-written policy in the Mispresentative and Exploitative domain. We provide further human evaluations in Appendix \Cref{section-human-evaluation}.}
\label{tab:safety-performance-taboola}
\end{table*}

The fully human-written policy remains strongest overall, but DPR narrows the gap substantially in several domains. Under majority voting, DPR reaches near parity on Misrepresentative, achieving 98\% of the human policy performance, and it comes within a small margin on Exploitative. At the same time, Finance Claims remains a clear outlier. This category relies heavily on organization-specific compliance language and fine-grained disclaimer patterns, which are difficult to recover from open web sources alone. These results highlight both the promise and limits of open-domain policy construction. DPR can quickly bootstrap useful draft sections from minimal input, especially in domains where conventions and edge cases are well represented in public guidance, while expert-written rules remain important for categories driven by proprietary standards.

Finally, DPR + Summary reduces the policy length but can trade off accuracy. The compressed policy retains most of the gains on Exploitative and Offensive but loses 1--2 F$_1$ on Finance Claims and Misrepresentative, suggesting that aggressive compression can remove rare qualifiers that the reader model uses for borderline decisions. We present qualitative examples of the generated rules and their relationship to expert policy sections in \Cref{section-qual-study}. We also provide a human evaluation of the generated rules in \Cref{section-human-evaluation}.

\section{Conclusion}
\label{section-discussion}
Deep Policy Research (DPR) shows that open-domain research agents can autonomously synthesize web-sourced information into effective safety policies. Across online text moderation and multimodal ad review, DPR outperforms definition-only prompts and example-based learning and often recovers nuanced edge cases and policy conventions, demonstrating the feasibility of drafting policies from high-level human guidance.

\bibliography{custom}

\clearpage
\appendix
\twocolumn[{%
 \centering
 \Large\bf Supplementary Material: Appendices \\ [20pt]
}]

\section{Implementation Details}
\label{section-further-impl-details}

\subsection{OpenAI Content Moderation}  
\label{section-further-impl-details-openai}
\paragraph{Evaluation} We select the five major domains (sexual, hate, violence, harassment, and self-harm) and use their official definition as the input for DPR\footnote{Accessed at \url{https://github.com/openai/moderation-api-release}.}, as presented in \Cref{example-policy-goals}. For each domain, we reserve five examples for validation and use the rest for testing. We perform evaluation separately for each domain using the in-domain undesired test examples as well as the safe test examples. To define the utility metric for the system-generated policy, we focus the evaluation scope on long-context evaluation with an off-the-shelf LLM. In other words, we directly provide the entire policy in the context of the LLM and prompt for the binary safe/unsafe judgment. 

\paragraph{DPR Implementation Details} We use the domain definitions provided in \Cref{example-policy-goals} as the policy specification $s$. Unless otherwise stated, we use Llama 3.3 70B Instruct \citep{DBLP:journals/corr/abs-2407-21783} as $\mathcal{M}$ and Google Search as $\mathcal{G}$. We run research for $k$ = 3 iterations and use $n$ = 20 for clustering. 

\subsection{In-House Multimodal Content Moderation}  
\label{section-further-impl-details-taboola}
\paragraph{Evaluation} We select four domains and provide a one-sentence summary as the input for DPR, as presented in \Cref{example-policy-goals}. We perform evaluation separately for each domain using the in-domain safe and unsafe examples. Similar to the previous experiment, we provide the entire policy in the context of an off-the-shelf large vision and language model\footnote{We used OpenAI \texttt{gpt-4o-2024-08-06}.} and prompt for the binary safe/unsafe judgment. We experiment in two inference settings: \textit{single sample (S.S)} and \textit{majority vote (M.V.)} with 10 responses sampled with temperature 0.8. 

\paragraph{DPR Implementation Details} We use Llama 3.3 70B Instruct as the research LLM. We keep all the hyperparameter settings consistent with the previous section and the only difference is the input domain specification. 

\subsection{Domain Specifications}
\label{section-dpr-detail-domain-specifications}

In \Cref{example-policy-goals}, we present all the domain definitions for all the experiments. These definitions serve as the seed domain specifications to run DPR. 

\subsection{Prompts}
\label{section-dpr-detail-prompts}

We provide full prompts for the DPR agent in \Cref{fig:prompt-generate-search-queries} through \Cref{fig:prompt-merge-section-titles}.

\clearpage
\begin{figure*}[ht!]
 \small
\centering
\begin{tabular}{p{0.95\linewidth}}
    \toprule
    \multicolumn{1}{c}{\texttt{Dataset: OpenAI Online Content Moderation}} \\
    \midrule
    \textit{Policy Domain: \textcolor{blue}{Sexual}} \\
    \textit{Definition: Content meant to arouse sexual excitement, such as the description of sexual activity, or that promotes sexual services (excluding sex education and wellness).} \\
    \vspace{0.1mm}
    \textit{Policy Domain: \textcolor{blue}{Hate}} \\
    \textit{Definition: Content that expresses, incites, or promotes hate based on race, gender, ethnicity, religion, nationality, sexual orientation, disability status, or caste.} \\
    \vspace{0.1mm}
    \textit{Policy Domain: \textcolor{blue}{Violence}} \\
    \textit{Definition: Content that promotes or glorifies violence or celebrates the suffering or humiliation of others.} \\
    \vspace{0.1mm}
    \textit{Policy Domain: \textcolor{blue}{Harassment}} \\
    \textit{Definition: Content that may be used to torment or annoy individuals in real life, or make harassment more likely to occur.} \\
    \vspace{0.1mm}
    \textit{Policy Domain: \textcolor{blue}{Self-Harm}} \\
    \textit{Definition: Content that promotes, encourages, or depicts acts of self-harm, such as suicide, cutting, and eating disorders.} \\
    \midrule
    \midrule
    \multicolumn{1}{c}{\texttt{Dataset: In-House Multi-Modal Advertisement Moderation}} \\
    \midrule
    \textit{Policy Domain: \textcolor{blue}{Offensive}} \\
    \textit{Definition: Advertisements that include graphic, gory, vulgar, or culturally insensitive content, or use imagery that shocks or offends viewers.} \\
    \vspace{0.1mm}
    \textit{Policy Domain: \textcolor{blue}{Exploitative}} \\
    \textit{Definition: Advertisements that use shocking, unsafe, or exploitative imagery—especially involving death, disasters, injuries, or sensitive groups.} \\
    \vspace{0.1mm}
    \textit{Policy Domain: \textcolor{blue}{Misrepresentative}} \\
    \textit{Definition: Advertisements that mislead users by implying guaranteed benefits, using false or outdated information, or showing unrelated or unrealistic content.} \\
    \vspace{0.1mm}
    \textit{Policy Domain: \textcolor{blue}{Problematic Finance Claims}} \\
    \textit{Definition: Advertisements that make extreme or misleading financial claims, guarantees, or unrealistic outcomes related to cost, savings, income, or risk.} \\
    \bottomrule  
\end{tabular}
\caption{Detailed specifications of the domains experimented in this paper. \textcolor{red}{These prompts were created solely for the purposes of this article and are provided for illustrative use only. They do not reflect official Taboola policy, which may be updated or revised over time.}}
\label{example-policy-goals}
\end{figure*}

\clearpage
\begin{figure*}[ht!]
\begin{tcolorbox}[colback=chocolate!5!white,colframe=chocolate!75!black,
                  title=Prompt for Generating Search Queries]
\begin{VerbatimWrap}
You are an expert in creating domain-specific knowledge bases. Given a research goal and a summary of the current knowledge datastore, you write a few querqies to Google for additional knowledge insufficiently covered by the current knowledge datastore.

Your research goal is: {research_goal}.

The current datastore summary: {current_datastore_summary}.

Write a list of Google queries that would find webpages that expand the coverage of the datastore. The queries should be in the form of a json list of strings, each string being a query. The queries should be relevant to the research goal and aim to cover gaps in the datastore. The queries should be specific. The queries can either directly ask for a specific information or ask for information from specific source types, which increases the likelihood of finding the right webpages.

Queries (in json list format):
\end{VerbatimWrap}
\end{tcolorbox}
\caption{\textbf{Prompt for generating web search queries.} The research agent uses it to identify missing coverage.}
\label{fig:prompt-generate-search-queries}
\end{figure*}

\begin{figure*}[ht!]
\begin{tcolorbox}[colback=chocolate!5!white,colframe=chocolate!75!black,
                  title=Prompt for Extracting Rules from a Webpage Chunk]
\begin{VerbatimWrap}
You are an expert in creating domain-specific knowledge bases. Given a research goal and content from Google, you summarize the relevant knowledge in the form of itemized rule.

Based on the following search results generate rules to represent the relevant knowledge.

Generate specific rules that:
1. Are directly extracted or derived from the search results provided.
2. Relevant to the research goal.
3. Cover different characteristics.
4. Are specific. Include any relevant nuances or edge cases mentioned.

VERY IMPORTANT: Your response MUST be a valid JSON array containing objects with these exact fields:
- "rule": the text of the rule
- "supporting_text": the exact quote from the passage that supports this rule

For example, your response should look exactly like this:
[
{
    "supporting_text": "Direct quote from the passage that supports the first rule",
    "rule": "Rule text goes here"
},
{
    "supporting_text": "Another direct quote that supports the second rule",
    "rule": "Another rule goes here"
}
]

Do not include any explanations, markdown formatting, or additional text before or after the JSON array.

### Your research goal:
{research_goal}

#### Search Result:
{webpage_chunk}

#### Rule (in json array format):
\end{VerbatimWrap}
\end{tcolorbox}
\caption{\textbf{Prompt for extracting rules from a webpage chunk.} The research agent uses it to generate new candidate rules.}
\label{fig:prompt-generate-rules-from-chunk}
\end{figure*}

\begin{figure*}[ht!]
\begin{tcolorbox}[colback=chocolate!5!white,colframe=chocolate!75!black,
                  title=Prompt for Scoring Rule Relevance]
\begin{VerbatimWrap}
You are an expert in creating domain-specific knowledge bases. Given a research goal and a piece of new knowledge you wanted to add to the knowledge datastore, represented as a rule, judge the relevance of the rule. The rule is relevant if it can be added to the knowledge base that answers the research goal. If the rule is only broadly related to the research goal, uninformative to answering the question posed in the research goal, or in the wrong format (e.g., asking for an action when the research is about definition), it is not relevant. Return your answer in a json dict with a single key 'relevance' and the value on a scale from 0 (irrelevant) to 10 (perfectly relevant).

##### Research Goal: {research_goal}.


New knowledge (represented as a rule): {rule_text}


##### Is this knowledge relevant enough? Directly write your evaluation in a json dict and do not write anything else:
\end{VerbatimWrap}
\end{tcolorbox}
\caption{\textbf{Prompt for scoring rule relevance.} The research agent uses it to filter candidate rules.}
\label{fig:prompt-critique-rule-relevance}
\end{figure*}

\begin{figure*}[ht!]
\begin{tcolorbox}[colback=chocolate!5!white,colframe=chocolate!75!black,
                  title=Prompt for Extracting Rule Keyphrases]
\begin{VerbatimWrap}
You are an expert in creating domain-specific knowledge bases. Given the domain description and an item in the knowledge base, write one keyphrase from the item. The keyphrase should identify the most salient information (concept or action) that distinguishs the item from the other items in the knowledge base. The information in domain description itself should not be in the keyphrase, because it is shared by all the items in the knowledge base.

##### Domain Description: {research_goal}.

##### Item: {rule_text}##### Keyphrase (a single phrase and nothing else):
\end{VerbatimWrap}
\end{tcolorbox}
\caption{\textbf{Prompt for extracting a keyphrase for each rule.} The research agent uses it during datastore indexing.}
\label{fig:prompt-extract-rule-keyphrase}
\end{figure*}

\begin{figure*}[ht!]
\begin{tcolorbox}[colback=chocolate!5!white,colframe=chocolate!75!black,
                  title=Prompt for Merging Similar Rules]
\begin{VerbatimWrap}
You are an expert in creating domain-specific knowledge bases. Given a domain description and some items from the knowledge base, combine similar items to make the list more concise. Output a list of json dicts, each dict corresponding to an item after your processing. Each dict must have two fields. The first field is "original_items", a list of items you choose to combine, exactly copied from the original items, and "new_item" is a string for the processed items. You should not combine items that are dissimilar. For the items you combine, make sure you cover all the information in the new item but do not write very long sentences. Instead, write a few shorter sentences to make the semantics clear.

##### Domain description: {research_goal}.

##### Original items:
{rule_text_list}

##### Processed items (output json array of dicts and nothing else):
\end{VerbatimWrap}
\end{tcolorbox}
\caption{\textbf{Prompt for merging similar rules.} The research agent uses it to consolidate extracted rules.}
\label{fig:prompt-merge-rules}
\end{figure*}

\begin{figure*}[ht!]
\begin{tcolorbox}[colback=chocolate!5!white,colframe=chocolate!75!black,
                  title=Prompt for Summarizing Section Rules]
\begin{VerbatimWrap}
You are an expert in creating domain-specific knowledge bases. Given a domain description and some similar items that form a single section, generate a short paragraph to summarize the topic of these items. The summary should serve as a good introduction to this section in the database. You should take the domain description into account and the summary should distinguish the items from the other potential sections under the same domain.

##### Domain description: {research_goal}.

##### Section Items:
{rule_text_list}

##### Section Summary (just output the summary text and nothing else):
\end{VerbatimWrap}
\end{tcolorbox}
\caption{\textbf{Prompt for summarizing a section of rules.} The research agent uses it to describe clustered sections.}
\label{fig:prompt-generate-section-summary}
\end{figure*}

\begin{figure*}[ht!]
\begin{tcolorbox}[colback=chocolate!5!white,colframe=chocolate!75!black,
                  title=Prompt for Titling Section Rules]
\begin{VerbatimWrap}
You are an expert in creating domain-specific knowledge bases. Given a domain description and some similar items that form a single section, as well as their associated keyphrases, generate a title for this section. You should take the domain description into account and the title should distinguish the items from the other potential sections under the same domain.

##### Domain description: {research_goal}.

##### Section Items:
{rule_text_list}

##### Keyphrases:
{keyphrases_list}

##### Section Title (just output the title text and nothing else):
\end{VerbatimWrap}
\end{tcolorbox}
\caption{\textbf{Prompt for titling a section of rules.} The research agent uses it to label clustered sections.}
\label{fig:prompt-generate-section-title}
\end{figure*}

\begin{figure*}[ht!]
\begin{tcolorbox}[colback=chocolate!5!white,colframe=chocolate!75!black,
                  title=Prompt for Merging Section Titles]
\begin{VerbatimWrap}
You are an expert in creating domain-specific knowledge bases. Given the domain definition a list of section titles, combine them into a more concise list by merging titles with the same meaning. Output a list of json dicts, each dict corresponding to an item after your processing. Each dict must have two fields. The first field is "original_titles", a list of items you choose to combine, exactly copied from the original items, and "new_title" is a string for the processed items. You should only combine titles that are similar enough. If the combined title is so general that it is equivalent to the domain description, do not combine.

##### Domain description: {research_goal}.

##### Existing titles:
{cluster_titles}

##### Combined section titles (in json list format and nothing else):
\end{VerbatimWrap}
\end{tcolorbox}
\caption{\textbf{Prompt for merging section titles.} The research agent uses it to reduce redundant titles.}
\label{fig:prompt-merge-section-titles}
\end{figure*}

\clearpage
\newpage
\section{Further Analyses}
\label{section-analyses-openai}

\subsection{Analyses on DPR Design}

\begin{figure*}[ht!]
 \centering
 \includegraphics[width=\linewidth]{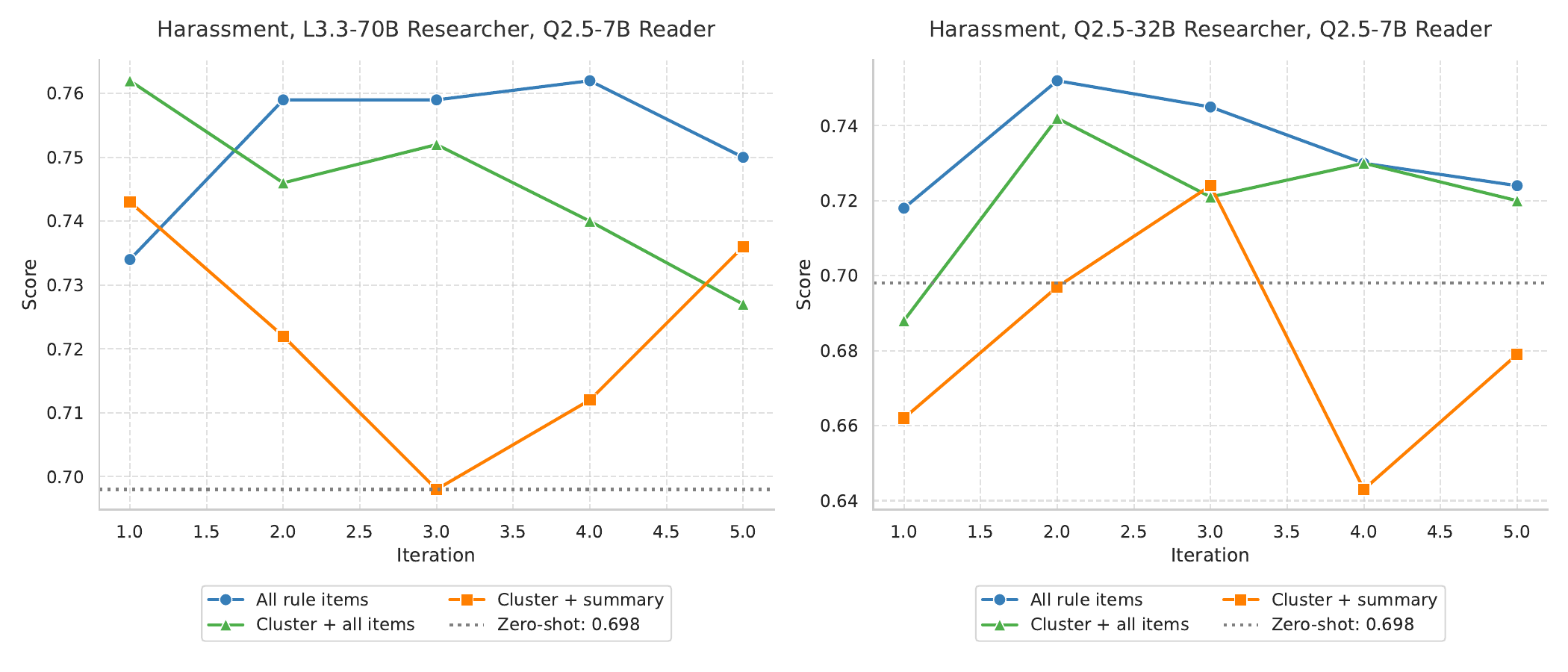}
 \caption{Performance vs. iteration in the Harassment domain with two different LLMs for DPR and Qwen2.5 7B Instruct as the reader.}
 \label{fig:convergence-dynamics-perf-qwen}
\end{figure*}

\begin{figure*}[ht!]
 \centering
 \includegraphics[width=\linewidth]{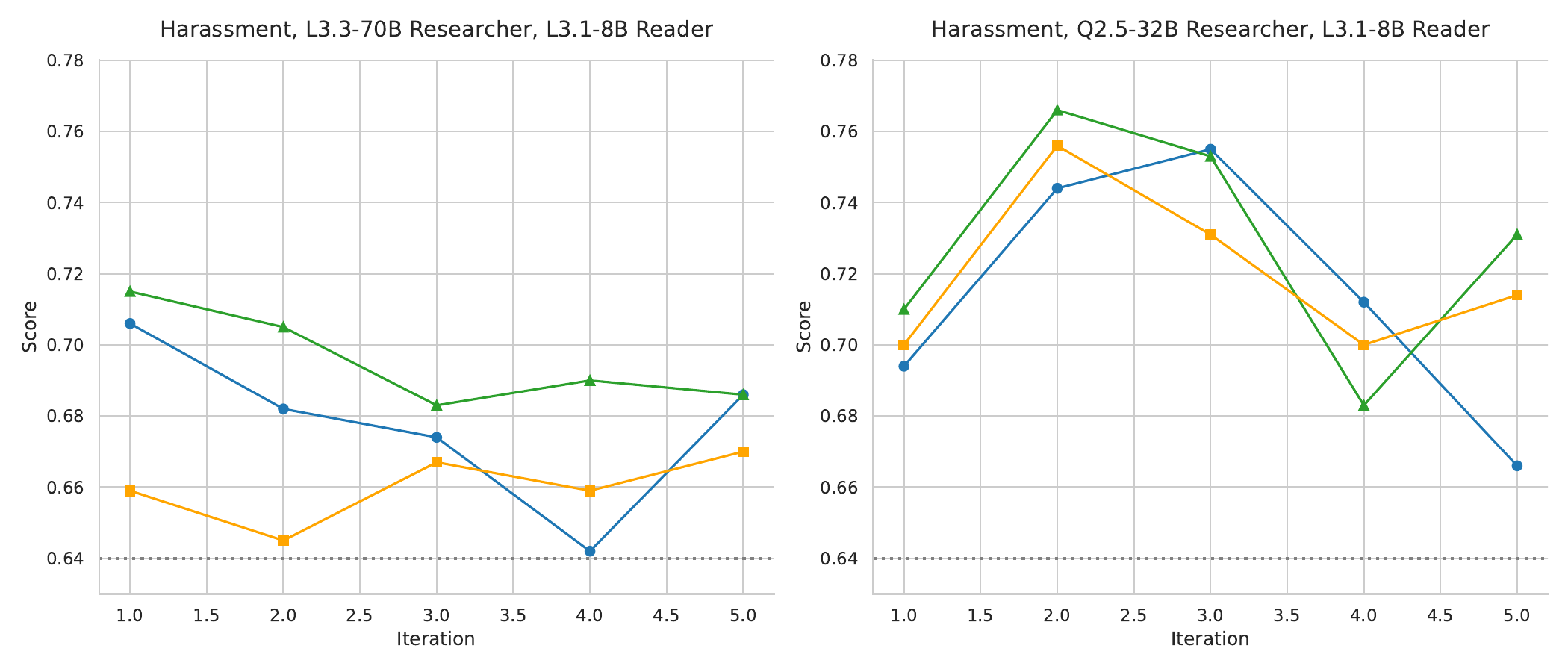}
 \caption{Performance vs. iteration in the Harassment domain with two different LLMs for DPR and Llama 3.1 8B Instruct as the reader.}
 \label{fig:convergence-dynamics-perf-llama}
\end{figure*}

\paragraph{Alternative Research Model.}
While the main experiments use Llama 3.3 70B for research, we
also ran the pipeline with a smaller Qwen2.5 32B \citep{yang2024qwen2}.  As shown in \Cref{fig:convergence-dynamics-perf-qwen} and \Cref{fig:convergence-dynamics-perf-llama}, we observe a similar F1 improvement over the baseline, suggesting the generality of the DPR framework. 

\begin{figure*}[ht!]
 \centering
 \includegraphics[width=\linewidth]{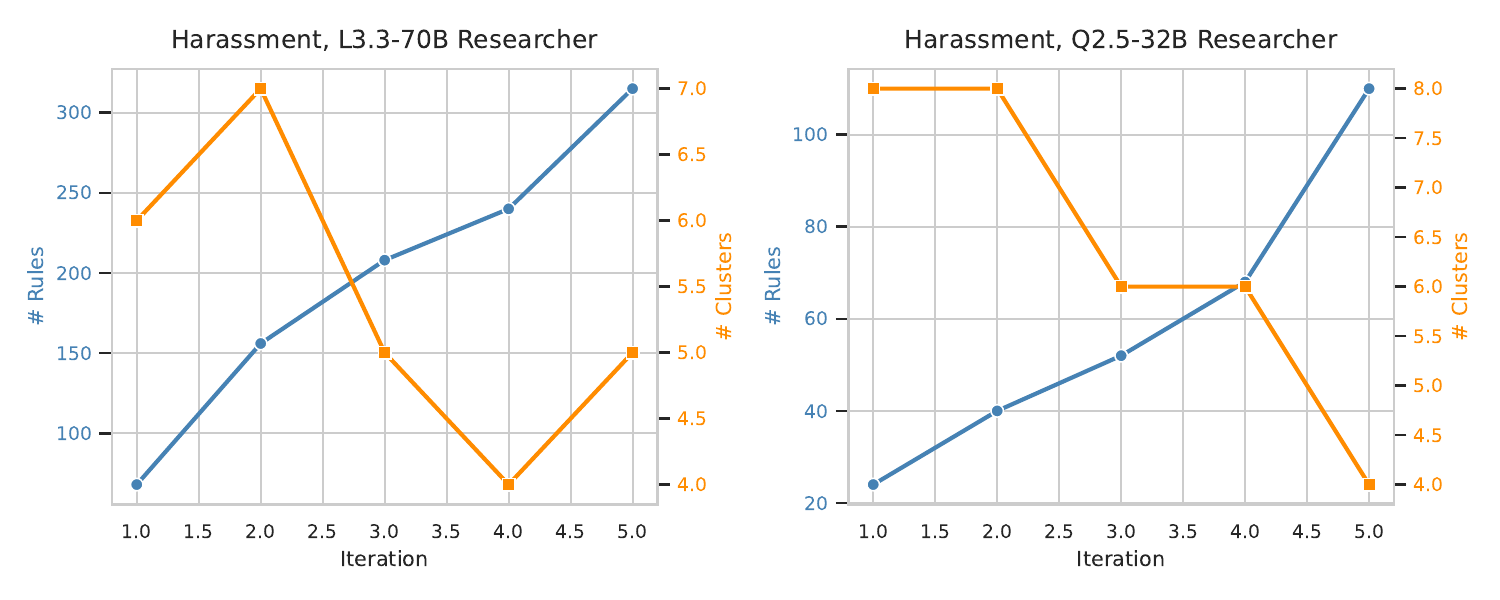}
 \caption{Rule count and cluster count vs. iteration in the Harassment domain for DPR initialized with two LLMs.}
 \label{fig:convergence-dynamics-counts}
\end{figure*}

\paragraph{Convergence Dynamics.}
\Cref{fig:convergence-dynamics-counts} tracks the number of unique
policy rules and the number of keyphrase clusters throughout five research iterations.  The curve exhibits a rapid expansion
followed by early stabilisation: over \(74\%\) of the final rules are
discovered in the first iteration, but only
\(8\%\) further clusters are added after the second.  This
plateau indicates diminishing returns beyond \(k=3\) iterations and
justifies our choice of the research budget.

\paragraph{Indexing Strategy.}
\Cref{fig:convergence-dynamics-perf-qwen} and \Cref{fig:convergence-dynamics-perf-llama} compare three ways of presenting
the same policy to the \emph{reader} LLM: (1) a full rule list with the section as the index,
(2) a keyphrase \emph{summary} of each cluster, and
(3) an ablated “flat” list without section headers.
Summaries cut the prompt length but come with a small
accuracy drop. The indexed rules and the flattened list have similar performance on Qwen2.5 7B Instruct. However, on Llama 3.1 8B Instruct, the former achieves the best performance. We hypothesize that Llama has a weaker long-context ability and thus can benefit more from more structured input formats.  

\begin{figure*}[ht!]
  \centering
  \includegraphics[width=\linewidth]{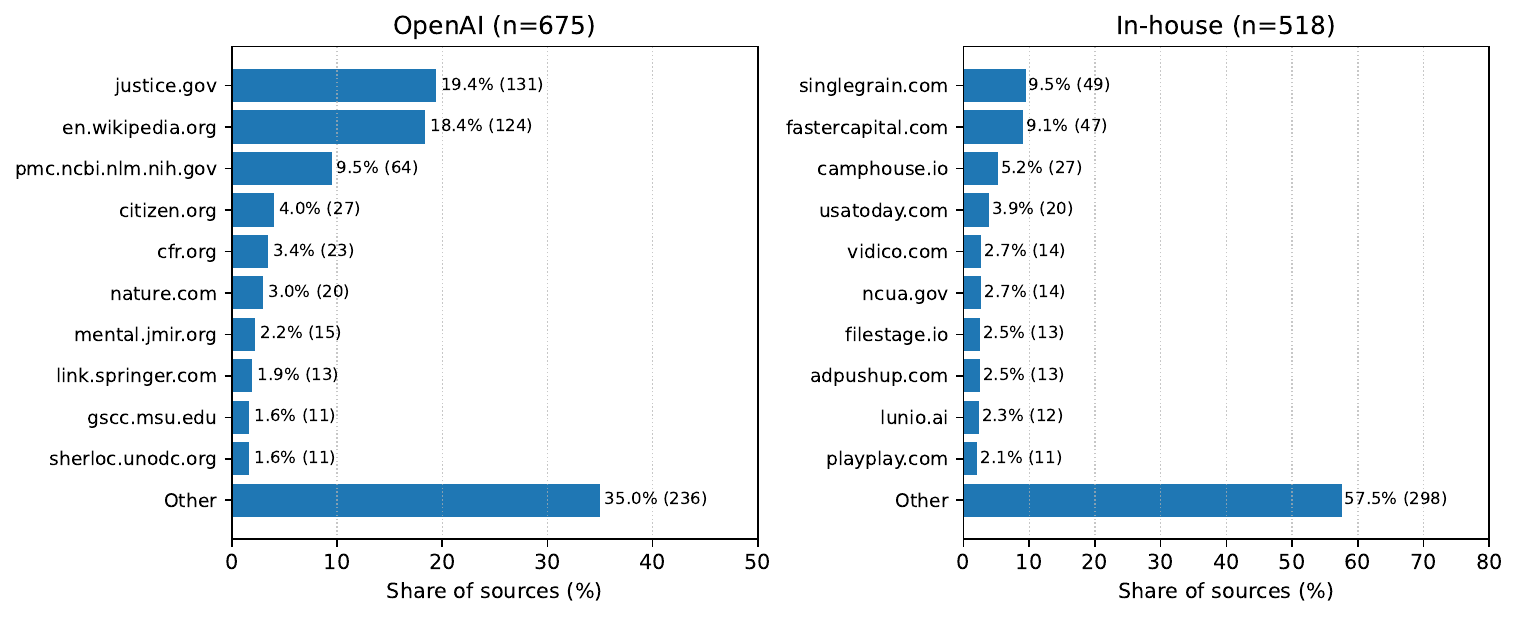}
  \caption{Domain distribution of sources leveraged by DPR for OpenAI and the in-house dataset.}
  \label{fig:domain-distribution}
\end{figure*}

\subsection{Domain Distribution}
\label{section-domain-distribution}

In \Cref{fig:domain-distribution}, we visualize the distribution of domains where DPR draws information from. We observe that DPR does not draw information exclusively from a single domain, nor does it directly copy policy from documents from model providers such as OpenAI that have similar purposes. Instead, it refers to the webpage from a diverse set of domain ranging from Wikipedia, scholarly articles, governmental documents, and various posts on the web. For the OpenAI content moderation, DPR relies more on formal articles to find strict definitions of the domains in interest. By contrast, for the in-house advertisement moderation task, DPR refers significantly more to marketing-related website to find more relevant information. 

\subsection{Qualitative Study}
\label{section-qual-study}

We provide qualitative studies of webpages, page excepts, and the corresponding DPR-generated rules in \Cref{tab:dpr_examples_openai} and  \Cref{tab:dpr_examples_taboola}. Overall, we observe that the model performs significant rephrasing and abstraction operations over the raw webpage information. In \Cref{fig:human_machine_rules_exploitative} and \Cref{fig:human_machine_rules_offensive}, we further compare DPR-generated policy with expert-written ones. We find that DPR often can provide rules that are close to expert policy in terms of both style and content. These results strengthen our belief in the potential of using agents to improve the automation in drafting policy documents.

\newpage
\begin{table*}[t]
\centering
\small
\setlength{\tabcolsep}{5pt}
\renewcommand{\arraystretch}{1.2}
\begin{tabularx}{\linewidth}{l l X X}
\toprule
\textbf{Domain} & \textbf{URL} & \textbf{Page Excerpt} & \textbf{DPR-Generated Rule} \\
\midrule

\multirow{2}{*}{Sexual} &
\href{https://thedailytexan.com/2018/04/19/study-shows-how-the-language-of-sexual-predators-in-chatrooms-differ/}{The Daily Texan (2018)} &
It found that sex offenders used significantly fewer first-person singular pronouns, such as ``I,'' ``me'' and ``my,'' and more second-person singular pronouns, such as ``you'' and ``your,'' compared to the decoys. &
Sensitive messages related to sexual content often use fewer first-person singular pronouns and more second-person singular pronouns. \\


\midrule

Hate &
\href{https://en.wikipedia.org/wiki/Online_hate_speech\#Characteristics_of_online_hate_speech}{Wikipedia: Online hate speech} &
Identity Tourism often leads to stereotyping, discrimination, and cultural appropriation. &
Hate messages may involve identity tourism, where a person pretends to be a member of another group. This can lead to stereotyping and discrimination. \\

\midrule

Violence &
\href{https://www.masteringculturaldifferences.com/blog/the-power-of-words-recognizing-the-alarming-normalization-of-violent-language}{Mastering Cultural Differences} &
The use of violent language and phrases like ``kill two birds with one stone,'' and ``bite the bullet,'' may seem innocuous on the surface. &
Sensitive messages related to violence may contain idiomatic phrases that appear innocuous but contribute to the desensitization of violence. \\

\midrule

Harassment &
\href{https://onlineharassmentfieldmanual.pen.org/defining-online-harassment-a-glossary-of-terms/}{PEN Field Manual Glossary} &
Dog Whistling: Using words or symbols with a double (or coded) meaning that is abusive or harmful, sometimes to signal a group of online abusers to attack a specific target. &
Harassment messages may involve dog whistling, using coded words or symbols with a double meaning to signal other abusers to attack the target. \\

\midrule

Self-Harm &
\href{https://www.bennet.senate.gov/2023/03/21/press-releases-id-2c176126-39db-424d-9eb3-2a7e0e699147/}{Sen. Bennet (Press Release)} &
In one case, researchers prompted My AI to instruct a child how to cover up a bruise ahead of a visit from Child Protective Services. &
Sensitive messages related to self-harm may include instructions on how to conceal physical harm or injuries. \\

\bottomrule
\end{tabularx}
\caption{Examples of DPR output on the OpenAI content moderation domains.}
\label{tab:dpr_examples_openai}
\end{table*}

\newpage
\begin{table*}[t]
\centering
\small
\setlength{\tabcolsep}{5pt}
\renewcommand{\arraystretch}{1.2}
\begin{tabularx}{\linewidth}{l l X X}
\toprule
\textbf{Domain} & \textbf{URL} & \textbf{Page Excerpt} & \textbf{DPR-Generated Rule} \\
\midrule

Misrepresentative &
\href{https://www.singlegrain.com/blog/a/misleading-statistics-in-advertising/}{SingleGrain: Misleading statistics} &
If a user finds an anti-aging skincare moisturizer that touts ``90\% of users experienced fewer wrinkles,'' this statistic will be enough for consumers to purchase the product. They won't realize or even care that it's a false advertising claim. &
Advertisements may use high percentages or statistics to create a misleading narrative, even if the claim is based on biased or incomplete data. \\

\midrule

Finance Claims &
\href{https://www.fma.govt.nz/consumer/investing/investing-basics/misleading-advertising-of-investment-products/}{FMA: Misleading advertising (investments)} &
What does the advertisement say (or not say) about the risk of the investment? Words such as ``secured'', ``certain'', ``stable'', ``guaranteed'' can give the impression that an investment is of low or no risk this can be misleading if the investment is not low risk. &
Advertisements for financial products should not use words that imply low or no risk, such as ``secured'', ``certain'', ``stable'', or ``guaranteed'', if the investment is not low risk. \\

\midrule

Exploitative &
\href{https://issuu.com/publicmediasolution/docs/shock_advertising_explained_with_examples}{Issuu: Shock advertising examples} &
PETA's ads often feature graphic imagery of animal cruelty, such as animals being slaughtered or abused in factory farms. &
Exploitative advertisements may use graphic imagery of animal cruelty, such as animals being slaughtered or abused, to promote a message or product. \\

\midrule

Offensive &
\href{https://www.lunio.ai/blog/top-controversial-ads}{Lunio: Controversial ads} &
Bumble released an ad campaign that sparked outrage online. As shown in the image, the billboard ad campaign included messages such as ``A vow of celibacy is not the answer'' and ``Thou shalt not give up on dating and become a nun''. &
Using religious references or imagery in a way that is perceived as mocking or disrespectful can be considered offensive. \\

\bottomrule
\end{tabularx}
\caption{Examples of DPR output on in-house advertisement policy domains.}
\label{tab:dpr_examples_taboola}
\end{table*}

\begin{figure*}[t!]
\small
\centering
\begin{tabular}{p{0.95\linewidth}}
\toprule
\multicolumn{1}{c}{\texttt{Human-written Rules and Machine-generated Rules}} \\
\midrule

\textit{Policy Domain: \textcolor{blue}{Exploitative}} \\

\vspace{0.6mm}
\textbf{Human-written rules.} \\

\textbf{5.1.} Must not use images of catastrophic scenes featuring violence or disaster. \\
\hspace*{1.2em}\textit{Sub-rules (5.1):} \\
\hspace*{2.4em}(a) Objects associated with death are permitted (e.g., Coffin, Urn, Tombstone). \\
\hspace*{2.4em}(b) Funerals, mourning and scenes associated with death are permitted. \\
\hspace*{2.4em}(c) Descriptions of catastrophic scenes are permitted in \textit{Titles} and \textit{Descriptions}. \\

\vspace{0.4mm}
\textbf{5.2.} Images depicting vehicle accidents must be carefully evaluated for both the visible vehicle damage and the condition of any individuals present. Images that strongly suggest a high-energy impact---with damage and contextual cues indicating a likelihood of life-threatening or severe injuries---must not be used. \\
\hspace*{1.2em}\textit{Sub-rules (5.2):} \\
\hspace*{2.4em}(a) Significant vehicle damage is acceptable if individuals exhibit only minor or superficial injuries and the overall context suggests minimal harm (e.g., calm behavior). \\
\hspace*{2.4em}(b) Cartoon images are permitted. \\

\vspace{0.4mm}
\textbf{5.5.} Must not use images of human remains being destroyed or in the process of being destroyed (e.g., coffin entering furnace). \\
\textbf{5.6.} Must not use images of known personalities who died in the last six months. \\
\textbf{5.7.} Must not use images of people performing physically dangerous tasks in an unsafe manner. \\
\textbf{5.8.} Must not use images of marijuana leaves or marijuana plants. \\
\textbf{5.9.} Must not target a specific ethnicity or religion with dating products or services. \\

\vspace{0.8mm}
\textbf{Machine-generated rules (with evidence).} \\

\vspace{0.4mm}
\textbf{Rule 1.} Exploitative advertisements may use images of fatal car crashes to promote their message, even if the circumstances of the crash are unrelated to the advertised issue. \\
\hspace*{1.2em}\textit{Closest human rule(s):} 5.2 \\
\hspace*{1.2em}\textit{Source:} \href{https://www.dailymail.co.uk/news/article-13205563/Diana-crash-used-vile-euthanasia-ad-campaign-Princesss-friends-condemn-cruel-exploitation-Paris-tragedy-photo-promotes-assisted-dying.html}{Daily Mail article} \\
\hspace*{1.2em}\textit{Excerpt:} The sickening campaign shows a picture of a mangled car in a tunnel with the caption: ``Diana. She did not choose her death... in 2024, we should have the choice''. \\

\vspace{0.6mm}
\textbf{Rule 2.} Exploitative advertisements may involve the post-mortem exploitation of a dead celebrity's image for financial gain. \\
\hspace*{1.2em}\textit{Closest human rule(s):} 5.6 \\
\hspace*{1.2em}\textit{Source:} \href{https://ccsenet.org/journal/index.php/ijms/article/view/25299}{IJMS article (CCSE)} \\
\hspace*{1.2em}\textit{Excerpt:} the post-mortem exploitation of a Deleb's image \\

\bottomrule
\end{tabular}
\caption{Visualization of human-written rules and aligned DPR-generated rules for the \textbf{Exploitative} domain. \textcolor{red}{These prompts were created solely for the purposes of this article and are provided for illustrative use only. They do not reflect official Taboola policy, which may be updated or revised over time.}}
\label{fig:human_machine_rules_exploitative}
\end{figure*}

\begin{figure*}[t!]
\small
\centering
\begin{tabular}{p{0.95\linewidth}}
\toprule
\multicolumn{1}{c}{\texttt{Human-written Rules and Machine-generated Rules}} \\
\midrule

\textit{Policy Domain: \textcolor{blue}{Offensive}} \\

\vspace{0.6mm}
\textbf{Human-written rules.} \\

\textbf{6.1.} Must not use images of a person with a visible exposed injury (e.g., on skin). \\
\hspace*{1.2em}\textit{Sub-rules (6.1):} \\
\hspace*{2.4em}(a) Images of bandaged or cast body parts are permitted. \\

\vspace{0.4mm}
\textbf{6.2.} Must not explicitly portray in images the inside of a human mouth. \\
\hspace*{1.2em}\textit{Sub-rules (6.2):} \\
\hspace*{2.4em}(a) Close ups of forced smiles showing teeth are prohibited. \\

\vspace{0.4mm}
\textbf{6.3.} Must not use gory or repulsive images. \\
\textbf{6.4.} Must not use images of a person with a needle or tube inserted into their skin or body. \\
\textbf{6.5.} Must not use an image of a person with a visible disease, condition or deformity that affects their appearance. \\
\hspace*{1.2em}\textit{Sub-rules (6.5):} \\
\hspace*{2.4em}(a) Images of medical procedures are permitted. \\

\vspace{0.4mm}
\textbf{6.6.} Must not use images of a dead body of a human or animal. \\
\textbf{6.7.} Must not feature a gun being pointed at the audience or to the edge of the frame. \\
\textbf{6.8.} Must not feature realistic ``fake teeth'', realistic veneers, realistic prosthetic teeth or realistic implants without any context. \\
\textbf{6.9.} Must not feature excessive amounts of realistic blood. \\
\textbf{6.10.} Must not use images of an unconscious or sedated person. \\
\textbf{6.11.} Must not use images of destroyed national symbols (e.g., images of burning cash, images of burning flags). \\
\textbf{6.12.} Must not use offensive language or symbols (e.g., ``Take a look at his repulsive home''). \\
\hspace*{1.2em}\textit{Sub-rules (6.12):} \\
\hspace*{2.4em}(a) Negative language or imagery is permitted. \\

\vspace{0.4mm}
\textbf{6.13.} Must not use language or images that are culturally or religiously insensitive. \\
\textbf{6.14.} Must not use vulgar or insensitive language or images. \\
\textbf{6.15.} Must not feature sensationalized stories about the Amish. \\
\textbf{6.16.} Must not make the focus of the story a personality as sexual orientation (e.g., ``Famous LGBT couples''). \\

\vspace{0.8mm}
\textbf{Machine-generated rules (with evidence).} \\

\vspace{0.4mm}
\textbf{Rule 1.} An advertisement can be considered \textit{Offensive} if it includes graphic or disturbing content, such as violence against animals. \\
\hspace*{1.2em}\textit{Closest human rule(s):} 6.3 \\
\hspace*{1.2em}\textit{Source:} \href{https://playplay.com/blog/controversial-commercials/}{PlayPlay: Controversial commercials} \\
\hspace*{1.2em}\textit{Excerpt:} The ad features scenes like firing gerbils from cannons and using a marching band as targets for hungry wolves. \\

\vspace{0.6mm}
\textbf{Rule 2.} Using religious references or imagery in a way that is perceived as mocking or disrespectful can be considered offensive. \\
\hspace*{1.2em}\textit{Closest human rule(s):} 6.13 \\
\hspace*{1.2em}\textit{Source:} \href{https://www.lunio.ai/blog/top-controversial-ads}{Lunio: Controversial ads} \\
\hspace*{1.2em}\textit{Excerpt:} Bumble released an ad campaign that sparked outrage online. As shown in the image, the billboard ad campaign included messages such as ``A vow of celibacy is not the answer'' and ``Thou shalt not give up on dating and become a nun''. \\

\vspace{0.6mm}
\textbf{Rule 3.} Offensive advertisements may include lewd or tasteless sexual references, obscenity, vulgarity, brutality, nudity, feces, profanity, or horrifying and repulsive images or words. \\
\hspace*{1.2em}\textit{Closest human rule(s):} 6.14 \\
\hspace*{1.2em}\textit{Source:} \href{https://en.wikipedia.org/wiki/Shock_advertising}{Wikipedia: Shock advertising} \\
\hspace*{1.2em}\textit{Excerpt:} They can include a disregard for tradition, law or practice (e.g., lewd or tasteless sexual references or obscenity), defiance of the social or moral code (e.g., vulgarity, brutality, nudity, feces, or profanity) or the display of images or words that are horrifying, terrifying, or repulsive (e.g., gruesome or revolting scenes, or violence). \\

\bottomrule
\end{tabular}
\caption{Visualization of human-written rules and aligned DPR-generated rules for the \textbf{Offensive} domain. \textcolor{red}{These prompts were created solely for the purposes of this article and are provided for illustrative use only. They do not reflect official Taboola policy, which may be updated or revised over time.}}
\label{fig:human_machine_rules_offensive}
\end{figure*}

\newpage
\clearpage

\section{Human Evaluation}
\label{section-human-evaluation}

We assess the perceived quality and internal applicability of DPR-generated rules by human annotation. The unit of annotation is a single policy rule as it appears in the consolidated, indexed policy for each domain. DPR merges newly generated rules with the previous iteration and organizes them via keyphrase-based clustering into an indexed document. We evaluate the rules in this final index.

\paragraph{Annotation Setup} The annotator group comprises three senior content reviewers with domain expertise in policy interpretation and enforcement within the in-house content ecosystem.

\paragraph{Rubric} For each domain and each rule, annotators provide the following ratings:
\begin{enumerate}
    \item Rule clarity \& actionability (linguistic): 
    \begin{itemize}
        \item 2 = clear and actionable, easy to apply; 
        \item 1 = mostly clear but some parts are vague; 
        \item 0 = too vague to be used in practice.
    \end{itemize}
    
    \item Rule domain relevance: 
    \begin{itemize}
        \item 2 = relevant to domain;
        \item 1 = vague (expected to appear in other domains as well);
        \item  0 = irrelevant.
    \end{itemize}
     
    \item Internal usability (in-house benchmark only):  relevance to the internal use case:
    \begin{itemize}
        \item 2 = directly relevant (should appear in internal policy); 
        \item 1 = potentially relevant ;
        \item 0 = probably irrelevant.
    \end{itemize}  
\end{enumerate}

\paragraph{Results} We report counts per label (0/1/2) and mean ± std per metric and domain (\Cref{tab:human-study-metrics-taboola} and \Cref{tab:human-study-metrics-oai}), and inter-annotator agreement using Fleiss’ $\kappa$ with observed agreement $\bar P $and chance agreement $P_e$ (\Cref{tab:human-study-iaa-taboola} and \Cref{tab:human-study-iaa-oai}). Across all evaluated domains, the human metrics in \Cref{tab:human-study-metrics-taboola,tab:human-study-metrics-oai} show that most rules are judged clear and actionable and domain-relevant, with the bulk of ratings concentrated at 1–2 on both scales. The small fraction of 0s typically reflects rules whose scope or preconditions require more context than the text provides. On the in-house only internal usability dimension, ratings are likewise skewed toward 1–2 but exhibit more spread, consistent with the fact that organizational constraints (e.g., business model, risk tolerance, and enforcement workflows) can make an otherwise well-formed rule only “potentially relevant.” Taken together, the counts and mean±std summaries indicate that the proposed policy synthesis generally yields rules that are understandable, scoped to their intended domain, and plausibly useful for internal moderation, with remaining gaps concentrated in conceptually diffuse or context-heavy areas (\Cref{tab:human-study-metrics-taboola,tab:human-study-metrics-oai}).

Inter-annotator agreement (IAA) in \Cref{tab:human-study-iaa-taboola,tab:human-study-iaa-oai} shows moderate to substantial Fleiss’ $\kappa$ overall, with a consistent pattern: $\kappa$ is typically higher for clarity/actionability than for domain relevance, and lowest for internal usability. This matches with intuition: clarity is more linguistic and thus easier to converge on; relevance requires topical judgment; and internal usability introduces organization-specific priors. Lower $\kappa$ pockets co-occur with categories that are inherently context dependent (e.g., content that hinges on intent, risk qualifiers, or external claims), signaling where additional guidance would most improve consistency. In short, the IAA results corroborate the descriptive metrics: the rules are broadly serviceable and legible to experts, and the remaining disagreement highlights precisely those areas where policy text benefits from tighter boundaries or richer adjudication cues (\Cref{tab:human-study-iaa-taboola,tab:human-study-iaa-oai}).

\begin{table*}[htbp]
\centering
\resizebox{\linewidth}{!}{
\begin{tabular}{l|cc|cc|cc}
\hline
\multirow{2}{*}{Domain} & \multicolumn{2}{c|}{Clarity} & \multicolumn{2}{c|}{Relevance} & \multicolumn{2}{c}{Internal Usability} \\
 & counts & mean$\pm$std & counts & mean$\pm$std & counts & mean$\pm$std \\
\hline
Exploitative      & 23/36/92  & 1.4570 $\pm$ 0.7460 & 20/49/83  & 1.4145 $\pm$ 0.7136 & 20/26/106 & 1.5658 $\pm$ 0.7157 \\
Finance           & 42/116/331 & 1.5910 $\pm$ 0.6437 & 43/77/369 & 1.6667 $\pm$ 0.6316 & 46/81/360 & 1.6865 $\pm$ 1.1256 \\
Misrepresentative & 47/47/59  & 1.0784 $\pm$ 0.8314 & 46/56/51  & 1.0327 $\pm$ 0.7982 & 53/51/49  & 0.9739 $\pm$ 0.8188 \\
Offensive         & 46/56/129 & 1.3593 $\pm$ 0.7945 & 38/78/115 & 1.3333 $\pm$ 0.7441 & 56/55/120 & 1.2771 $\pm$ 0.8295 \\
\hline
\end{tabular}
}
\caption{Human evaluation on in-house test domains: counts of ratings (0/1/2) and mean$\pm$std for each metric. }
\label{tab:human-study-metrics-taboola}
\end{table*}

\begin{table*}[htbp]
\centering
\begin{tabular}{l|cc|cc}
\hline
\multirow{2}{*}{Domain} & \multicolumn{2}{c|}{Clarity} & \multicolumn{2}{c}{Relevance} \\
 & counts & mean$\pm$std & counts & mean$\pm$std \\
\hline
Harassment & 7/28/172  & 1.7971 $\pm$ 0.4801 & 9/26/172  & 1.7874 $\pm$ 0.5055 \\
Hate       & 25/35/231 & 1.7079 $\pm$ 0.6164 & 27/36/228 & 1.6907 $\pm$ 0.6329 \\
Self-Harm  & 44/30/51  & 1.0560 $\pm$ 0.8735 & 37/37/51  & 1.1120 $\pm$ 0.8349 \\
Sexual     & 14/16/45  & 1.4133 $\pm$ 0.7900 & 14/20/41  & 1.3600 $\pm$ 0.7822 \\
Violence   & 17/38/71  & 1.4286 $\pm$ 0.7203 & 22/47/57  & 1.2778 $\pm$ 0.7445 \\
\hline
\end{tabular}
\caption{Human evaluation on OpenAI domains: counts of ratings (0/1/2) and mean$\pm$std for each metric. }
\label{tab:human-study-metrics-oai}
\end{table*}

\begin{table*}[htbp]
\centering
\resizebox{\linewidth}{!}{
\begin{tabular}{l|ccc|ccc|ccc}
\hline
\multirow{2}{*}{Domain} & \multicolumn{3}{c|}{Clarity} & \multicolumn{3}{c|}{Relevance} & \multicolumn{3}{c}{Internal Usability} \\
 & $\kappa$ & $\bar{P}$ & $P_e$ & $\kappa$ & $\bar{P}$ & $P_e$ & $\kappa$ & $\bar{P}$ & $P_e$ \\
\hline
Exploitative      & 0.6069 & 0.7843 & 0.4513 & 0.4146 & 0.6601 & 0.4194 & 0.5802 & 0.8039 & 0.5329 \\
Finance           & 0.0377 & 0.5399 & 0.5218 & -0.1354 & 0.5481 & 0.6020 & -0.1280 & 0.5297 & 0.5830 \\
Misrepresentative & 0.7830 & 0.8562 & 0.3374 & 0.6853 & 0.7908 & 0.3355 & 0.8038 & 0.8693 & 0.3337 \\
Offensive         & 0.7504 & 0.8528 & 0.4103 & 0.5679 & 0.7359 & 0.3889 & 0.8098 & 0.8831 & 0.3853 \\

\hline
\end{tabular}

}
\caption{Inter-annotator agreement on in-house test domains per metric (Fleiss' $\kappa$), observed agreement $\bar{P}$, and chance agreement $P_e$.}
\label{tab:human-study-iaa-taboola}
\end{table*}

\begin{table*}[htbp]
\centering
\begin{tabular}{l|ccc|ccc}
\hline
\multirow{2}{*}{Domain} & \multicolumn{3}{c|}{Clarity} & \multicolumn{3}{c}{Relevance} \\
 & $\kappa$ & $\bar{P}$ & $P_e$ & $\kappa$ & $\bar{P}$ & $P_e$ \\
\hline
Harassment & 0.6337 & 0.8937 & 0.7099 & 0.7021 & 0.9130 & 0.7081 \\
Hate       & 0.6840 & 0.8900 & 0.6520 & 0.7344 & 0.9038 & 0.6378 \\
Self-Harm  & 0.4888 & 0.6667 & 0.3480 & 0.4454 & 0.6349 & 0.3417 \\
Sexual     & 0.7618 & 0.8667 & 0.4404 & 0.5072 & 0.7067 & 0.4048 \\
Violence   & 0.4186 & 0.6667 & 0.4267 & 0.2643 & 0.5397 & 0.3743 \\
\hline
\end{tabular}
\caption{Inter-annotator agreement on OpenAI test domains per metric (Fleiss' $\kappa$), observed agreement $\bar{P}$, and chance agreement $P_e$.}
\label{tab:human-study-iaa-oai}
\end{table*}

\end{document}